\documentclass[sigconf,screen,nonacm]{acmmm_template/acmart}

\renewcommand\footnotetextcopyrightpermission[1]{}
\usepackage{graphicx}
\usepackage{booktabs}
\usepackage{subcaption}
\usepackage{hyperref}
\usepackage{orcidlink}
\usepackage[ruled,vlined]{algorithm2e}
\usepackage{xcolor}
\usepackage[table]{xcolor}
\usepackage{enumitem}

\makeatletter

\DeclareRobustCommand\onedot{\futurelet\@let@token\@onedot}
\def\@onedot{\ifx\@let@token.\else.\null\fi\xspace}
\def\eg{\emph{e.g}\onedot} 
\def\ie{\emph{i.e}\onedot}

\makeatother

\begin{document}


\title{Mobile GUI Agent Privacy Personalization with Trajectory Induced Preference Optimization}

\author{Zhixin Lin$^{1}$, Jungang Li$^{2,3}$, Dongliang Xu$^{1\dagger}$, Shidong Pan$^{4}$\\
Yibo Shi$^{5}$, Yuchi Liu$^{6}$, Yuecong Min$^{7}$, Yue Yao$^{1\dagger}$}

\affiliation{
  \vspace{4pt}
  \institution{
    $^{1}$Shandong University \quad
    $^{2}$The Hong Kong University of Science and Technology (Guangzhou) \\
    $^{3}$The Hong Kong University of Science and Technology \quad
    $^{4}$New York University \quad
    $^{5}$Xi'an Jiaotong University \quad
    $^{6}$Australian National University \quad
    $^{7}$Institute of Computing Technology, Chinese Academy of Sciences
  }
  \country{}
}

\begin{abstract}


Mobile GUI agents powered by Multimodal Large Language Models (MLLMs) can execute complex tasks on mobile devices.
Despite this progress, most existing systems still optimize task success or efficiency, neglecting users' privacy personalization.
In this paper, we study the often-overlooked agent personalization.
We observe that personalization can induce systematic structural heterogeneity in execution trajectories. 
For example, privacy-first users often prefer protective actions (\eg, refusing permissions, logging out, minimizing exposure), leading to logically different execution trajectories from utility-first users. 
This variable-length and structurally different execution trajectory makes standard preference optimization unstable and less informative. 
To address this, we propose Trajectory Induced Preference Optimization (TIPO), which uses preference-intensity weighting to emphasize key privacy-related steps and padding gating to suppress alignment noise. 
Results on our Privacy Preference Dataset show that TIPO improves persona alignment and distinction while preserving strong task executability, achieving 65.60\% SR, 46.22 Compliance, and 66.67\% PD, outperforming existing optimization methods across various GUI tasks. The code and dataset will be publicly released at \url{https://github.com/Zhixin-L/TIPO}.

\end{abstract}


\maketitle
\begingroup
\renewcommand\thefootnote{}
\footnotetext{$\dagger$ Corresponding authors.}
\endgroup
\section{Introduction}
\label{sec:intro}

%

\begin{figure}[t]
    \centering
    \includegraphics[width=\linewidth]{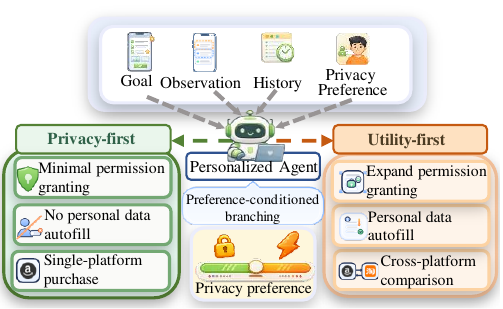}
    \caption{Illustration of personalized trajectory selection for smartphone GUI agents. Given the same task goal, observation history, and user portrait, different privacy personas can induce different preference-conditioned branches, leading to distinct execution trajectories.}
    \label{fig:1_illustration}

\end{figure}
%
%

Unlike traditional voice assistants, which only answer questions in natural language, mobile GUI agents can directly operate apps to complete real user instructions, such as searching for information across apps, sending messages, booking tickets, navigating maps, adjusting system settings, managing emails, and completing shopping or service workflows~\cite{liu2025llm,zhang2024large,shi2026androtmem}. 
In recent years, Multimodal Large Language Models (MLLMs) have enabled mobile GUI agents to perform complex tasks on mobile devices~\cite{li2025mobileuse,tang2025survey}. 
The advancement of mobile GUI agents further improved task completion rate on these realistic smartphone tasks~\cite{rawles2024androidworld,jiang2025appagentx,xu2026mobile}.
They are moving beyond proof-of-concept demonstrations toward practical assistants that can act on behalf of users in daily routines.

However, from an end-user's perspective, \textit{task completion} alone does not necessarily imply \textit{user satisfaction}~\cite{siro2022understanding,kiseleva2016predicting}. 
Users care not only about whether a task is completed, but also about \textit{how} it is completed and what risks are incurred along the way, \eg, privacy exposure. For the same task, there are often multiple feasible execution trajectories, and different users may prefer different trade-offs between utility and privacy risk. Shown in Fig.~\ref{fig:1_illustration}, privacy-sensitive users tend to behave more conservatively during task execution, for example, by reading privacy policies carefully, granting only necessary permissions, disabling personalized tracking, and logging out of accounts to reduce unnecessary exposure. ~\cite{pan2024read,liu2022protecting,hutton2023exploring} In contrast, utility-oriented users are more likely to accept default settings and follow more direct, higher-utility paths, even at the cost of greater privacy exposure, in order to reduce interaction friction and accomplish tasks more efficiently.

Most existing research and systems optimize \emph{Success Rate} or interaction efficiency, by assuming that each task has a single optimal trajectory, and ignoring trajectory induced
preferences by user preference (\eg, the privacy preferences)~\cite{nguyen2025gui,zhang2025appagent}. 
This gap significantly limits the real-world user experience of Mobile GUI agents.
In this paper, we thereby focus on a practical yet often overlooked direction: \emph{personalized operation trajectory selection based on user privacy preferences} in the Mobile GUI agent context. 
We define the personalized trajectory selection as follows: for multiple trajectories that can complete the same task, the agent should select one that best aligns with the preferences implied by the user preference. 
Specifically, we focus on \textit{privacy} preference as a representative and high-stakes user preference that critically affects the trustworthiness of Mobile GUI agents. 
In this work, privacy is operationalized as a user's preference regarding how much personal information is disclosed, retained, tracked, or exposed during task execution. 
This preference is reflected in concrete behavioral choices, such as whether to grant optional permissions, accept personalized tracking, remain logged in, or clear traces after task completion. 
Inspired by Westin’s categorization of privacy attitudes~\cite{elueze2018privacy} and further extending it, privacy preference allows us to clearly identify which steps are necessary and which steps should be avoided, making it an ideal entry point for evaluating ``process personalization.'' 
To support this setting, we build a new \emph{Privacy Preference} dataset with paired trajectories under different privacy personas.

An intuitive solution for personalized trajectory selection is preference-aligned training on an MLLM-based Mobile GUI agent, where a policy learns from paired feedback between \emph{chosen} (more preference-consistent) and \emph{rejected} (less consistent) trajectories. 
However, we find that existing preference optimization methods, such as  Direct Preference Optimization (DPO)~\cite{rafailov2023direct}, are poorly matched to this problem. 
Trajectories from different privacy preferences are often structurally heterogeneous and length-mismatched.
For example, for the same task goal, one trajectory may directly complete the target action, while another may additionally adjust privacy-related settings, deny unnecessary permissions, or clear traces after completion.
As a result, the training signal is diluted, and gradients can be dominated by padding. The model may learn how to match the padded format rather than learning the true privacy-related differences. Moreover, privacy-relevant actions are often sparse but critical within a long trajectory; standard DPO is not sufficiently sensitive to these key steps, leading to unstable optimization and limited gains. 

To address the intrinsic heterogeneity, we propose \emph{Trajectory Induced Preference Optimization (TIPO)}, a preference optimization method for structurally heterogeneous trajectories. TIPO improves learning in two ways: it uses \emph{preference-intensity weighting} to emphasize persona-relevant steps, and \emph{padding gating} to suppress noise from alignment placeholders. Together, these designs make preference optimization more suitable for variable-length trajectory pairs.
Results show that TIPO achieves the best overall performance among compared methods, reaching 65.60\% in SR, 42.85 in PAS-S, 46.22 in Compliance, and 66.6\% in PD, while preserving strong task executability. 
These results indicate that TIPO not only maintains task success but also more effectively aligns the generated trajectories with the target privacy persona and strengthens persona distinction across diverse mobile tasks. 

Further ablations show that preference-intensity weighting improves learning on privacy-critical steps, raising Compliance from 31.94 (DPO) to 38.93, while the full model further increases it to 46.22; similarly, PD improves from 59.26\% to 62.96\% and finally to 66.67\%, demonstrating that padding gating and preference-intensity weighting are complementary and jointly yield the most stable and consistent gains.

Further ablations indicate that preference-intensity weighting strengthens learning on critical privacy steps, while padding gating effectively suppresses unstable updates caused by alignment noise; their combination yields the most stable and consistent improvements. 
These results demonstrate that user preference can induce structural trajectory differences in Mobile GUI agents, and that preference optimization mechanisms designed for such differences can effectively address the problem.

In summary, our main contributions are:
\begin{enumerate} [leftmargin=*, topsep=0pt]
  \item We define mobile \textbf{GUI agent personalization} as a trajectory selection task, where the goal is not only to complete the task but also to choose a persona-consistent trajectory under the same task objective.
  \item We build the \textbf{privacy preference} dataset, providing multi-trajectory annotations for the same task goal under different privacy preferences, characterized by variable length and structural heterogeneity.
  \item We propose the \textbf{Trajectory
Induced Preference Optimization} (TIPO), which stabilizes preference optimization for variable-length, structurally heterogeneous trajectory feedback through preference-intensity weighting and padding gating, leading to significant improvements in experiments.
\end{enumerate}

\section{Related Work}
\subsection{Preference Optimization for Alignment}
Recent alignment research has increasingly shifted from Reinforcement Learning from Human Feedback (RLHF) toward simpler offline preference optimization objectives ~\cite{wang2024comprehensive,winata2025preference,liu2025survey}. Direct Preference Optimization (DPO)~\cite{rafailov2023direct} is a representative example, showing that preference alignment can be achieved with a direct classification-style objective without explicit reward modeling or reinforcement learning. Subsequent work has explored several variants of this paradigm, including reference-free or simplified objectives such as ORPO~\cite{hong2024orpo} and SimPO~\cite{meng2024simpo}, contrastive formulations such as CPO~\cite{xu2024contrastive}, and broader theoretical perspectives such as IPO~\cite{azar2024general}. More recent studies further revisit reference mismatch, token-level weighting~\cite{zeng2024token}, and explicit preference objectives~\cite{hu2025explicit}, suggesting that preference optimization remains an active and evolving direction.

However, most of these methods are developed for settings in which the compared outputs are relatively homogeneous, such as two responses to the same prompt. Our setting is different. In Mobile GUI agents, user preferences can induce variable-length and structurally heterogeneous trajectories. A preferred trajectory may differ not only in local action choice, but also in whether certain steps are inserted, skipped, or reorganized. This makes standard response-level preference optimization less suitable, since the compared units are no longer naturally aligned. Our TIPO is therefore designed for trajectory-level preference learning under structural heterogeneity.
\begin{figure}[t]
    \centering
    \includegraphics[width=\linewidth]{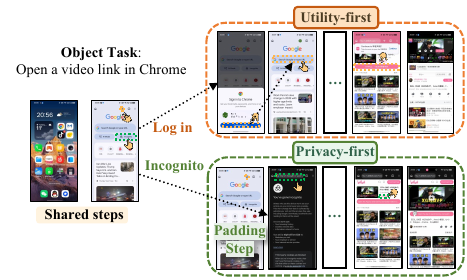}
    \caption{A showcase of persona-induced trajectory divergence. Under the same task goal, Utility-first and Privacy-first share the same initial steps but diverge at a privacy-sensitive decision point, resulting different executable trajectories.}
    \label{fig:6_case}
\end{figure}
\subsection{Personalization and User Modeling}
Personalized alignment aims to move beyond population-level behavior and adapt models to users' preferences, histories, and decision styles~\cite{liu2025survey2,guan2025survey,xie2025survey}. Existing work studies this problem through personalized preference learning, progressive adaptation, and benchmark construction. Representative examples include P-RLHF~\cite{li2024personalized}, which introduces personalized preference learning, PROPER~\cite{zhang2025proper}, which formulates personalization as progressive refinement, and recent benchmarks such as PersonaLens~\cite{zhao2025personalens} and Persona2Web~\cite{kim2026persona2web}, which make personalized behavior increasingly measurable in conversational and agent settings.

However, most existing personalization frameworks are developed for dialogue systems or web agents, where personalization is mainly reflected in response content or high-level decisions. In Mobile GUI agents, user preferences directly affect execution behavior, including permission handling, account states, privacy exposure, and risk-related action choices. As a result, personalization reshapes not only what the agent does, but also how the entire action trajectory is organized. This makes Mobile GUI personalization a problem of trajectory-level structural variation, highlighting the need for a trajectory-centric framework such as ours.

\subsection{Mobile GUI Agents and Mobile Privacy}
Recent progress in multimodal large language models has led to the rapid development of Mobile GUI agents and mobile interaction benchmarks.~\cite{shi2025towards,wang2024gui,nguyen2025gui} Early systems such as AppAgent~\cite{zhang2025appagent} and AppAgent-v2~\cite{li2024appagent} demonstrate the feasibility of autonomous mobile app operation, while benchmarks such as AndroidWorld~\cite{rawles2024androidworld}, GUIOdyssey~\cite{lu2025guiodyssey}, and SPA-Bench~\cite{chen2024spa} make evaluation more realistic and systematic. More recent agent systems, including UI-TARS~\cite{qin2025ui} and Mobile-Agent-v3~\cite{ye2025mobile}, further push this direction toward stronger grounding, longer-horizon execution, and more practical deployment.

Recent studies have also started to examine privacy in mobile agent settings, shifting attention from general task execution to privacy-related risks and protections. However, existing work still mainly focuses on task success, privacy awareness~\cite{lin2025mind}, or information protection~\cite{zhao2026anonymization}. We study how user-specific privacy preferences reshape the execution trajectory itself under the same task goal. This makes privacy preference a trajectory-selection problem rather than only a detection or protection problem, motivating our trajectory-centric method.
\section{Problem Definition}

We define Mobile GUI agents personalization as a trajectory selection task. Specifically, given the same task and initial UI state, different user preferences may induce different preferred execution strategies, leading to trajectories with systematically different structures and lengths. 
In this work, we focus on a high-stakes preference \emph{privacy preference}, and aim to generate trajectories that are more consistent with the target privacy preference while preserving task feasibility.
Figure~\ref{fig:6_case} showcase a visualized example.
%

%
We define the agent input as \(x=(g,o,h,p)\), where \(g\) is the task goal, \(o\) is the current UI observation, \(h\) denotes a bounded interaction history, and \(p\) denotes the privacy persona. To enable a controlled study, we instantiate \(p\) in a simplified binary form, with two representative profiles: Privacy-first and Utility-first. Given \(x\), the agent generates a step sequence \(y=(y_1,\dots,y_{|y|})\), where each \(y_t\) denotes a specific action (\eg, ``tap search'' or ``open the JD app''). Since privacy preference may introduce additional defensive actions or suppress utility-oriented ones, trajectory lengths are often different across persona branches.

We construct persona-conditioned preference training samples as triplets $(x, y^{+}, y^{-})$, where $y^{+}$ denotes the trajectory that is more aligned with the privacy persona specified in $x$, and $y^{-}$ denotes a less aligned alternative under the same task context. This follows the standard preference-pair formulation used in DPO. Let $\pi_{\theta}(y \mid x)$ denote the agent policy parameterized by $\theta$. The objective is to encourage the policy to assign higher probability to the persona-aligned trajectory than to the less aligned one under the same context, \ie,
\[
\pi_{\theta}(y^{+} \mid x) > \pi_{\theta}(y^{-} \mid x).
\]
Importantly, this preference relation is defined over privacy alignment rather than task success, as both trajectories are assumed to be feasible solutions to the task.

\section{Method}

\subsection{Preliminaries}




\textbf{Direct Preference Optimization (DPO).}
We adopt DPO as the base preference optimization framework. Given a persona-conditioned preference pair $(x, y^{+}, y^{-})$, where $y^{+}$ and $y^{-}$ denote the preferred and less preferred trajectories under the same context $x$, DPO encourages the policy model $\pi_{\theta}$ to assign a higher relative likelihood to $y^{+}$ than to $y^{-}$ with respect to a fixed reference policy $\pi_{\mathrm{ref}}$.

In standard DPO, this preference is defined at the trajectory level. Specifically, the sequence-level preference score can be written as
\[
\begin{aligned}
z(\theta)
&=
\beta \Big[
\big(\log \pi_{\theta}(y^{+}\mid x)-\log \pi_{\theta}(y^{-}\mid x)\big) \\
&\qquad\quad
-\big(\log \pi_{\mathrm{ref}}(y^{+}\mid x)-\log \pi_{\mathrm{ref}}(y^{-}\mid x)\big)
\Big].
\end{aligned}
\]
where $\beta>0$ controls the sharpness of the preference signal.

The DPO objective is then written as
\[
L_{\mathrm{DPO}}
=
\mathbb{E}_{(x,y^{+},y^{-})}
\big[
\mathrm{softplus}(-z(\theta))
\big].
\]
\textbf{Limitations of DPO and motivation of our methods.}
DPO implicitly treats aligned positions in a preference pair as equally valid comparison units. This assumption becomes problematic in our setting, where persona-induced trajectories may differ substantially in both structure and length. After alignment, some positions correspond to genuine persona-related decisions, while others are introduced only for padding, leading to semantic noise and making uniform supervision less suitable for trajectory preference learning.

As a result, standard DPO faces two limitations in our setting:
\begin{itemize} [leftmargin = *]
    \item \textbf{Padding alignment introduces semantic placeholder noise.} To compare variable-length trajectories, the chosen and rejected branches must be aligned to a common length. However, the resulting \texttt{no\_action} placeholders do not carry genuine preference information, yet they still enter the loss calculation.
    \item \textbf{Uniform treatment of aligned positions ignores step importance.} In standard DPO, all aligned positions are treated as equally informative training units. Consequently, persona-critical steps may be diluted by a large number of neutral or placeholder positions.
\end{itemize}

As a result, standard DPO becomes less effective on variable-length, structurally heterogeneous trajectory pairs. This limitation motivates TIPO, which improves preference learning by emphasizing persona-critical steps and suppressing alignment-induced noise.
\begin{figure}[t]
    \centering
    \includegraphics[width=\linewidth]{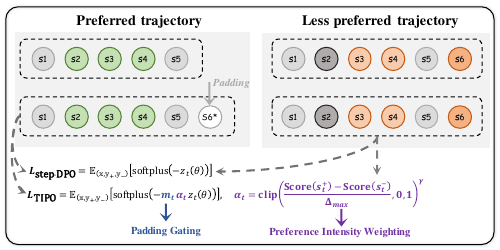}
    \caption{Comparison between step-DPO and TIPO on aligned trajectory pairs. While step-DPO treats positions uniformly, including alignment-induced placeholders, TIPO highlights persona-critical steps via preference-intensity weighting and reduces placeholder noise through padding gating.}
    \label{fig:2_algorithm}
    \vspace{-2mm}
\end{figure}
\subsection{TIPO}

Since our method operates on aligned trajectory steps, we further decompose this sequence-level comparison into step-wise preference signals. Let $x_t$ denote the step-level planning context at aligned step $t$, including the current task state and the relevant interaction history up to that step. We define
\[
\begin{aligned}
z_t(\theta)
&=\beta\Big[
\big(\log \pi_\theta(y_t^{+}\mid x_t)-\log \pi_\theta(y_t^{-}\mid x_t)\big) \\
&\qquad\quad
-\big(\log \pi_{\mathrm{ref}}(y_t^{+}\mid x_t)-\log \pi_{\mathrm{ref}}(y_t^{-}\mid x_t)\big)
\Big].
\end{aligned}
\]
This step-wise form is a decomposition of the original trajectory-level DPO objective tailored to our aligned-action setting, rather than the standard DPO formulation itself.
\[
L_{\mathrm{step\text{-}DPO}}
=
- \log \sigma\!\left(z_t(\theta)\right).
\]
As shown in Fig.~\ref{fig:2_algorithm}, TIPO improves step-level preference optimization from two complementary aspects: it emphasizes persona-critical aligned positions through preference intensity weighting, and suppresses noise through a padding gating mechanism.

\textbf{Preference intensity weighting.}
Standard DPO treats all aligned positions uniformly during optimization, which may dilute persona-critical decisions in long trajectories. In our setting, however, different aligned positions contribute unequally to privacy preference expression: some steps (\eg, denying permissions or disabling tracking) are highly informative for persona discrimination, while many neutral steps mainly serve task execution and carry little preference information. To address this issue, we assign each aligned step pair a preference intensity weight \(\alpha_t\), so that persona-relevant positions contribute more strongly to optimization.

Specifically, for each aligned step pair \((s_t^+, s_t^-)\), we first compute a persona-aware score difference
\[
\Delta s_t
=
\operatorname{Score}(s_t^+)
-
\operatorname{Score}(s_t^-),
\]
where \(\operatorname{Score}(\cdot)\) is derived from a rule-based action scoring scheme with LLM assistance. Concretely, our annotation protocol specifies preference-related action categories and their corresponding scores, while the LLM is used only to assist semantic normalization and resolve cases where surface forms differ but the underlying action intent is equivalent. The detailed scoring rules are provided in the appendix.

We then map \(\Delta s_t\) to a normalized step weight
\[
\alpha_t
=
\operatorname{clip}\left(\frac{\Delta s_t}{\Delta_{\max}}, 0, 1\right)^\gamma,
\]
where \(\Delta_{\max}\) is the maximum score difference used for normalization, and \(\gamma \ge 0\) controls the sharpness of the mapping. A larger \(\gamma\) assigns relatively higher weights to steps with stronger persona relevance.
\begin{figure*}[t]
    \centering
    \includegraphics[width=0.95\linewidth]{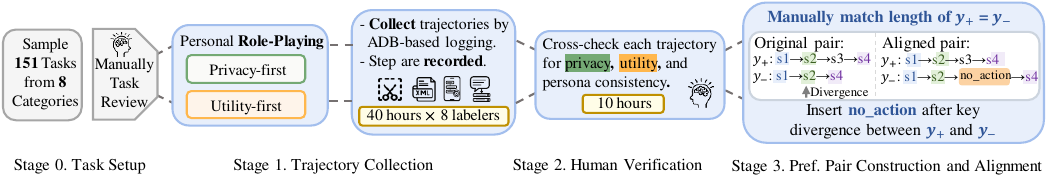}
    \caption{An overview of the Privacy Preference Dataset construction pipeline. For each task, we collect paired trajectories under both Privacy-first and Utility-first personas, followed by human verification and pairwise alignment to construct trajectory-level preference pairs.}
    \label{fig:3_data_construction}
\end{figure*}
\textbf{Padding gating mechanism.}
Although preference-intensity weighting highlights persona-relevant steps, aligned trajectory pairs still contain \texttt{no\_action} placeholders introduced by variable-length alignment. These positions do not carry genuine preference information and may interfere with optimization if treated the same as valid semantic steps. To suppress such alignment-induced noise, we introduce a padding gate \(m_t\):
\[
m_t=
\begin{cases}
0, & \text{if the chosen is } \texttt{no\_action},\\
1, & \text{otherwise}.
\end{cases}
\]
The gated weighted preference score is then defined as
\[
\hat{z}^{(\mathrm{gate})}_t(\theta)=m_t\,\alpha_t\,z_t(\theta).
\]
Finally, the TIPO objective is
\[
L_{\mathrm{TIPO}}
=
\mathbb{E}_{(x_t, y_t^+, y_t^-)}
\left[
\mathrm{softplus}\!\left(-\hat{z}^{(\mathrm{gate})}_t(\theta)\right)
\right].
\]

\section{Experiments and Results}
\label{sec:experiments}

\subsection{Dataset Construction}

To study privacy preference driven personalized trajectory selection, we build the Privacy Preference dataset using the pipeline shown in Fig.~\ref{fig:3_data_construction}, which combines real persona role-playing with on-device human trajectory collection. Inspired by Westin’s categorization of privacy attitudes~\cite{elueze2018privacy}, we instantiate privacy preference in a simplified binary setting with two personas: Privacy-first and Utility-first. 
This design captures the utility-privacy trade-off central to our problem, while remaining extensible to finer-grained privacy personas in future work.

The dataset covers eight high-frequency task categories (Shopping, Payment, Browsing, Food Delivery, Sharing, Account, Backup, and Reservations). In total, it contains \textbf{151} task instances. Each task is annotated with paired trajectories under \textbf{Privacy-first} and \textbf{Utility-first} personas, resulting in \textbf{302} trajectories and approximately \textbf{8.2k} annotated steps. Each task provides a clear natural-language goal instruction (\eg, ``\textit{Open a YouTube link \dots in the Edge browser}''). For representation, we treat a step as the minimal interaction unit and represent each execution as a trajectory of steps. 

\begin{table*}
\centering
\caption{Comparison of TIPO against baseline methods on the Privacy Preference Dataset. Results are reported in terms of task success rate (SR), persona adherence (PAS-S/PAS-U), compliance, and persona distinction (PD) under Privacy-first (P-f) and Utility-first (U-f) conditions. Otherwise indicated, the higher is better for values in this table. 
}
\label{table1.tex}
\begin{tabular}{c|ccc|cc|cc|cc|c} 
\hline
                              & \multicolumn{3}{c|}{\textbf{SR}}                                                                                                                                   & \multicolumn{2}{c|}{\textbf{PAS-S}}                                                                     & \multicolumn{2}{c|}{\textbf{PAS-U}}                                                                     & \multicolumn{2}{c|}{\textbf{Compliance}}                                                                & \multicolumn{1}{c}{\textbf{PD}}                      \\ 
\hline
\textbf{Methods}              & \textit{P-f}                                         & \textit{U-f}                                         & \textit{Overall}                                     & \textit{P-f}                                       & \textit{U-f ↓}                                     & \textit{P-f↓}                                      & \textit{U-f}                                       & \textit{Comp}                                      & \textit{Non-comp↓}                                 & /                                                     \\ 
\hline
Frozen                        & 44.08\%                                              & 34.50\%                                              & 39.29\%                                              & 16.13                                              & 78.57                                              & 88.89                                              & \textbf{\textcolor[rgb]{0.29,0.529,0.498}{65.31}}  & \textbf{\textcolor[rgb]{0.659,0.804,0.776}{40.72}} & 83.73                                              & 48.15\%                                               \\
SFT                           & \textbf{\textcolor[rgb]{0.659,0.804,0.776}{70.21\%}} & 60.13\%                                              & 65.17\%                                              & \textbf{\textcolor[rgb]{0.659,0.804,0.776}{35.60}} & \textbf{\textcolor[rgb]{0.659,0.804,0.776}{32.38}} & 37.04                                              & 38.12                                              & 36.86                                              & 34.71                                              & 59.26\%                                               \\
DPO~\cite{rafailov2023direct} & \textbf{\textcolor[rgb]{0.29,0.529,0.498}{70.31\%}}  & 60.37\%                                              & 65.34\%                                              & 28.20                                              & 46.67                                              & \textbf{\textcolor[rgb]{0.29,0.529,0.498}{14.81}}  & 35.67                                              & 31.94                                              & \textbf{\textcolor[rgb]{0.659,0.804,0.776}{30.74}} & 59.26\%                                               \\
ORPO~\cite{hong2024orpo}      & 69.65\%                                              & 61.31\%                                              & \textbf{\textcolor[rgb]{0.659,0.804,0.776}{65.48\%}} & 17.26                                              & 46.67                                              & 54.32                                              & 43.90                                              & 30.58                                              & 50.50                                              & 48.15\%                                               \\
CPO~\cite{xu2024contrastive}  & 68.63\%                                              & \textbf{\textcolor[rgb]{0.659,0.804,0.776}{61.36\%}} & 65.00\%                                              & 19.73                                              & 39.52                                              & 61.73                                              & 48.82                                              & 34.28                                              & 50.63                                              & 48.15\%                                               \\
SimPO~\cite{meng2024simpo}    & 67.60\%                                              & 60.62\%                                              & 64.11\%                                              & 25.07                                              & 37.14                                              & 51.85                                              & 46.69                                              & 35.88                                              & 44.50                                              & \textbf{\textcolor[rgb]{0.659,0.804,0.776}{62.96\%}}  \\
IPO~\cite{azar2024general}    & 70.04\%                                              & 60.84\%                                              & 65.44\%                                              & 29.02                                              & 39.52                                              & 37.04                                              & 42.09                                              & 35.56                                              & 38.28                                              & \textbf{\textcolor[rgb]{0.659,0.804,0.776}{62.96\%}}  \\
\textbf{Ours}                 & 69.08\%                                              & \textbf{\textcolor[rgb]{0.29,0.529,0.498}{62.11\%}}  & \textbf{\textcolor[rgb]{0.29,0.529,0.498}{65.60\%}}  & \textbf{\textcolor[rgb]{0.29,0.529,0.498}{42.85}}  & \textbf{\textcolor[rgb]{0.29,0.529,0.498}{15.71}}  & \textbf{\textcolor[rgb]{0.659,0.804,0.776}{29.63}} & \textbf{\textcolor[rgb]{0.659,0.804,0.776}{49.59}} & \textbf{\textcolor[rgb]{0.29,0.529,0.498}{46.22}}  & \textbf{\textcolor[rgb]{0.29,0.529,0.498}{22.67}}  & \textbf{\textcolor[rgb]{0.29,0.529,0.498}{66.67\%}}   \\
\hline
\end{tabular}
\end{table*}

Each step records (1) an executable structured action represented as \texttt{action\_type (arguments)}, for example \texttt{tap (x=129, y=138)}.
The complete action space and parameter specifications are provided in the Appendix.
(2) the visual observation of the current screen; (3) the corresponding UI state in XML format, and when necessary, a reasoning text explaining preference-driven decisions (\eg, a Privacy-first persona enables incognito mode when opening a browser); and (4) a semantic description of the action (\eg, ``tap the search bar to enter a query''). This multi-view logging of ``action parameters + screenshot + UI structure + semantic annotation'' makes the dataset suitable for supervised learning and also facilitates manually auditing the authenticity and executability of trajectories.
%
%
%
%

To ensure data quality and reproducibility, the dataset was annotated by eight annotators, each contributing about 40 hours, for a total of approximately 320 annotation hours. Before formal annotation, all annotators completed a warm-up stage, including a shared pilot annotation and two additional days of trial annotation, to familiarize themselves with the annotation interface, persona-specific rules, and trajectory recording requirements.
We used Android Debug Bridge (ADB) to capture on-device screenshots together with executable action traces, so that each step is grounded in verifiable UI evidence.
Because the collection process may involve privacy-sensitive scenarios, annotators were strictly prohibited from using any real personal information. 
All collected data were inspected before inclusion, and only samples verified to contain no private information were retained.
For quality assurance, each trajectory was cross-checked by a second annotator, who verified both persona-rule consistency and semantic consistency between the recorded actions and their corresponding step descriptions.
When disagreements arose, they were resolved through discussion under the unified annotation guidelines before the trajectory was finalized.
In addition, for every task, the paired Privacy-first and Utility-first trajectories were required to contain at least one key persona-differentiating action.

\textbf{Preference pair construction and alignment.}
For the same task instance, we manually role-play two user preferences, Privacy-first and Utility-first, to execute the task and obtain two executable semantic trajectories under the same goal constraint. For a given preference $p$, we define the trajectory consistent with the persona as the preferred trajectory $y^{+}$, and the other branch as the dispreferred trajectory $y^{-}$. Therefore, the preference pair $(x, y^{+}, y^{-})$ arises naturally during annotation.

Importantly, in mobile tasks, preference differences often induce structural trajectory heterogeneity. For example, Privacy-first trajectories may include additional defensive actions, such as reducing exposure or logging out, which makes the preferred and dispreferred branches systematically different in both trajectory structure and length. As a result, for a preference pair \((x, y^{+}, y^{-})\), the two trajectories are often not directly comparable at the sequence level, \ie, \(|y^{+}| \neq |y^{-}|\). Since DPO-style training compares preferred and dispreferred trajectories under a unified sequence dimension, we first align each pair to a common length
\(T=\max(|y^{+}|,|y^{-}|)\).
To do so, we introduce a semantic placeholder action \texttt{no\_action} to fill missing positions in the shorter branch. This placeholder is defined at the trajectory level rather than as a tokenizer-level padding token, and indicates that no corresponding semantic step exists at that position. We use an LLM to assist divergence-point identification between the two trajectories, followed by manual verification, and insert \texttt{no\_action} into the shorter trajectory until both branches have equal length. This yields aligned trajectory pairs that remain semantically interpretable and can be used for subsequent preference optimization.

\begin{figure}[t]
    \centering
    \includegraphics[width=\linewidth]{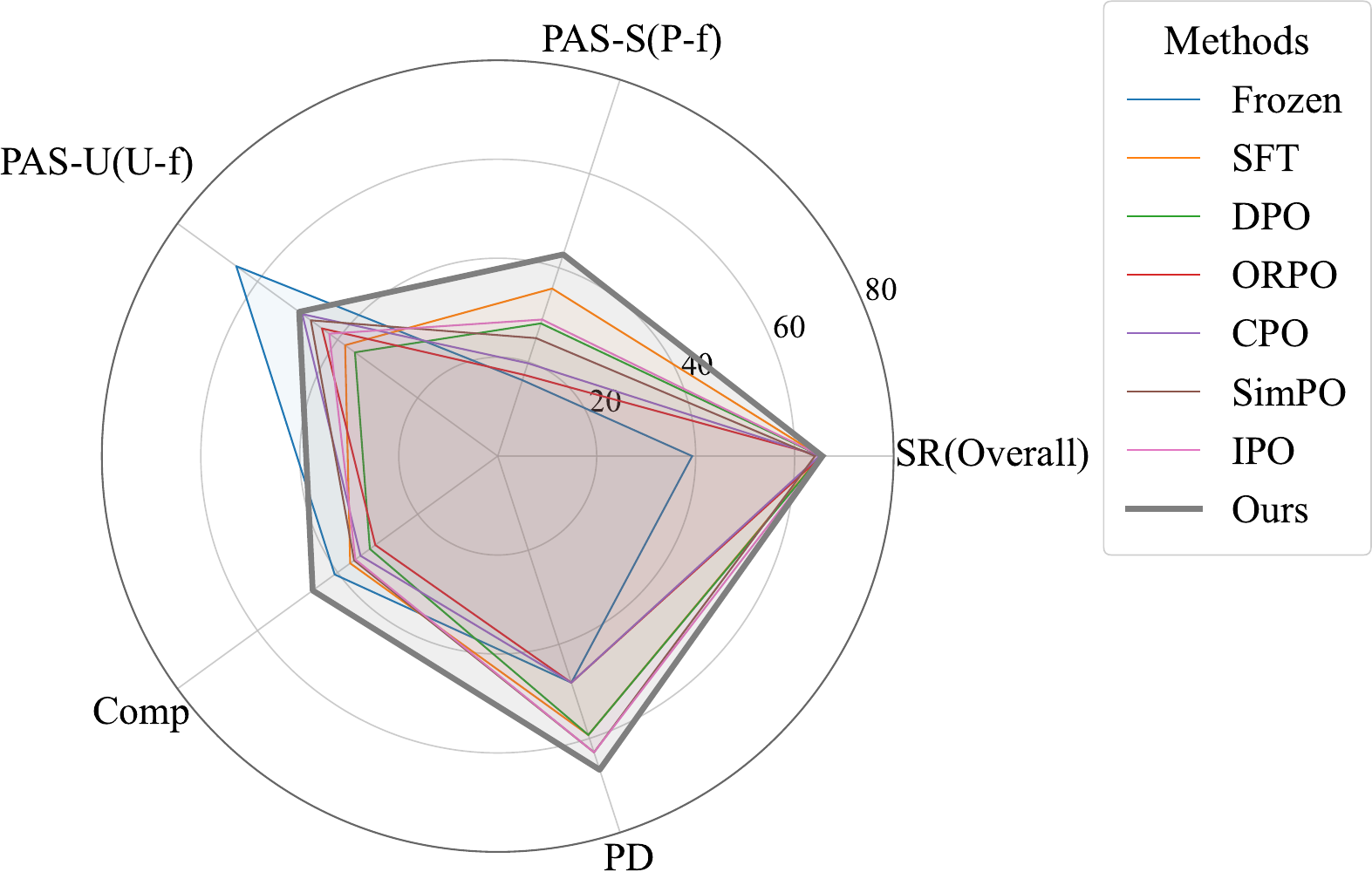}
    \caption{Radar-chart comparison of different methods on representative \emph{higher-is-better} metrics, including Overall SR, PAS-S under Privacy-first, PAS-U under Utility-first, Compliance, and PD. The figure provides a holistic view of the trade-off between task executability, persona adherence, and persona distinction across methods.}
    \label{fig:4_radar}
\end{figure}
\subsection{Experimental Settings}
To systematically evaluate the effectiveness of the proposed \emph{Trajectory Induced Preference Optimization (TIPO)} for personalized trajectory selection in Mobile GUI agents, we conduct experiments from three perspectives: task executability, persona-level preference consistency, and persona distinction ability. Under a unified experimental protocol, we compare TIPO with several representative preference optimization baselines.

%

%

\textbf{Task setting.} 
We consider the following task setting: given the same task goal and the same initial UI state, the agent is required to generate an execution trajectory that is consistent with the target user's privacy persona. 
We adopt the standard \emph{Planner--Executor} paradigm. Since our focus is personalized selection at the trajectory level, both training and evaluation are centered on the semantic trajectories generated by the Planner.
In addition, we adopt the same binary privacy persona setting introduced in the dataset construction section, namely Privacy-first (P-f) and Utility-first (U-f). We evaluate whether the model can generate persona-consistent trajectories under the same task goal and initial UI state while preserving task feasibility.

\textbf{Dataset and experimental setup.}
We conduct experiments on the Privacy Preference Dataset introduced above. To avoid information leakage caused by different branches of the same task goal appearing in both training and test stages, we adopt a task-level split to construct the training, validation, and test sets. Specifically, trajectories associated with different personas under the same task instance are always placed in the same data subset and never distributed across different subsets.

We use Qwen2.5VL-3B~\cite{bai2025qwen2} as the backbone model in all experiments, as it provides strong multimodal understanding for mobile UI scenarios while maintaining a moderate model size that is practical for controlled comparison across multiple preference-optimization methods.
Except for the Frozen method, all trainable methods are supervised fine-tuned (SFT) on the same training data to acquire basic UI understanding and semantic step generation capabilities, and are then further optimized on the same preference-pair data for preference alignment. To verify the effectiveness of our method, we compare TIPO with several representative baselines, including Frozen, SFT, DPO~\cite{rafailov2023direct}, ORPO~\cite{hong2024orpo}, IPO~\cite{azar2024general}, SimPO~\cite{meng2024simpo}, and CPO~\cite{xu2024contrastive}. Except for differences in the optimization objective itself, all methods are trained and evaluated under the same data split, unified input protocol, and consistent training settings to ensure a fair comparison.

\textbf{Evaluation Metrics and Measurement Protocol.}
To comprehensively evaluate model performance on the personalized trajectory selection task, we measure results from three perspectives: task performance, persona adherence, and persona distinction.

\textbf{ (1) Task Performance.}
We use \textbf{Step Success Rate (SR)} as the basic task performance metric, and report results separately for Privacy-first, Utility-first, and Overall. This metric measures the degree of step-level alignment between the generated trajectory and the reference trajectory. Specifically, we adopt a two-stage matching protocol. We first determine whether the action types are consistent. If the action types match, we further assess whether the generated step and the reference step are semantically equivalent. The final SR is averaged at the trajectory level and then aggregated over the test set.

\textbf{ (2) Persona Adherence.}
To evaluate whether a generated trajectory is consistent with the target privacy persona, we measure persona adherence using two behavior dimensions: security-oriented behaviors and utility-oriented behaviors. Specifically, we denote by $S(\tau)$ the operations that reflect defensive and privacy-preserving tendencies, and by $U(\tau)$ the operations that are utility-oriented but may increase privacy exposure risk.

For the model-generated trajectory $\tau_{\mathrm{infer}}$ and the reference trajectory $\tau_{\mathrm{gt}}$, we compute normalized ratios along both dimensions, denoted as PAS-S and PAS-U, respectively. For Privacy-first users, the desired behavior is to preserve security-oriented actions while suppressing utility-oriented but privacy-risky actions; therefore, higher PAS-S and lower PAS-U indicate better persona adherence. For Utility-first users, the desired behavior is the opposite: the model is expected to stay closer to the utility-oriented reference trajectory, so lower PAS-S and higher PAS-U are preferred.

To provide a more compact summary, we further report two aggregated metrics. Compliance is defined as the average of the two persona-consistent directions, i.e., PAS-S under Privacy-first and PAS-U under Utility-first, where higher values are better. Non-compliance is defined as the average of the two persona-inconsistent directions, i.e., PAS-U under Privacy-first and PAS-S under Utility-first, where lower values are better. Together, these metrics evaluate not only whether the model reproduces desired persona-consistent behaviors, but also whether it avoids persona-inconsistent ones.

\begin{table}
\centering
\caption{Ablation results of TIPO on the Privacy Preference Dataset. We compare different component variants in terms of overall task success (SR), persona adherence (PAS-S/PAS-U), compliance, and persona distinction (PD).}
\label{table2_single-column.tex}
\setlength{\tabcolsep}{4.5pt}
\begin{tabular}{cccccc} 
\hline
                 & \textbf{SR}                                          & \textbf{PAS-S}                                     & \textbf{PAS-U}                                     & \textbf{Compliance}                                & \textbf{PD}                                           \\ 
\hline
\textbf{Methods} & \textit{Overall}                                     & \textit{P-f}                                       & \textit{U-f}                                       & \textit{Comp}                                      & /                                                     \\ 
\hline
DPO              & 65.34\%                                              & 28.20                                              & 35.67                                              & 31.94                                              & 59.26\%                                               \\
w/o pw           & 65.26\%                                              & 29.84                                              & \textbf{\textcolor[rgb]{0.659,0.804,0.776}{41.20}} & 35.52                                              & 59.26\%                                               \\
w/o pg           & \textbf{\textcolor[rgb]{0.659,0.804,0.776}{65.59\%}} & \textbf{\textcolor[rgb]{0.659,0.804,0.776}{40.54}} & 37.32                                              & \textcolor[rgb]{0.659,0.804,0.776}{\textbf{38.93}} & \textcolor[rgb]{0.659,0.804,0.776}{\textbf{62.96\%}}  \\
\textbf{Ours}    & \textbf{\textcolor[rgb]{0.29,0.529,0.498}{65.60\%}}  & \textbf{\textcolor[rgb]{0.29,0.529,0.498}{42.85}}  & \textbf{\textcolor[rgb]{0.29,0.529,0.498}{49.59}}  & \textbf{\textcolor[rgb]{0.29,0.529,0.498}{46.22}}  & \textcolor[rgb]{0.29,0.529,0.498}{\textbf{66.67}\%}   \\
\hline
\end{tabular}
\vspace{-2mm}
\end{table}
\textbf{ (3) Persona Distinction.}
Consistency under a single persona is not sufficient to demonstrate true personalization ability. We therefore introduce Persona Distinction (PD) to evaluate whether, for the same task, the model can generate two trajectories that remain logically valid but exhibit clearly different preference orientations when only the privacy preference is changed. Specifically, for each task instance, we generate two trajectories under the Privacy-first and Utility-first settings, respectively, and compare their relative behaviors in terms of security-oriented and utility-oriented operations. If the results generated satisfy the predefined persona differentiation criteria, the case is counted as a success. The final PD score is averaged over all test tasks.

Since persona adherence and persona distinction involve trajectory-level preference judgments, some metrics require both predefined behavioral rules and semantic-level equivalence judgments. To ensure evaluation consistency, we apply the same evaluation prompts and decision protocol to all methods, and manually inspect a subset of samples to verify the reliability of the automatic evaluation results. Unless otherwise specified, all reported results are obtained under the same evaluation setting.

\section{Results}

%
\begin{figure}[t]
    \centering
    \includegraphics[width=\linewidth]{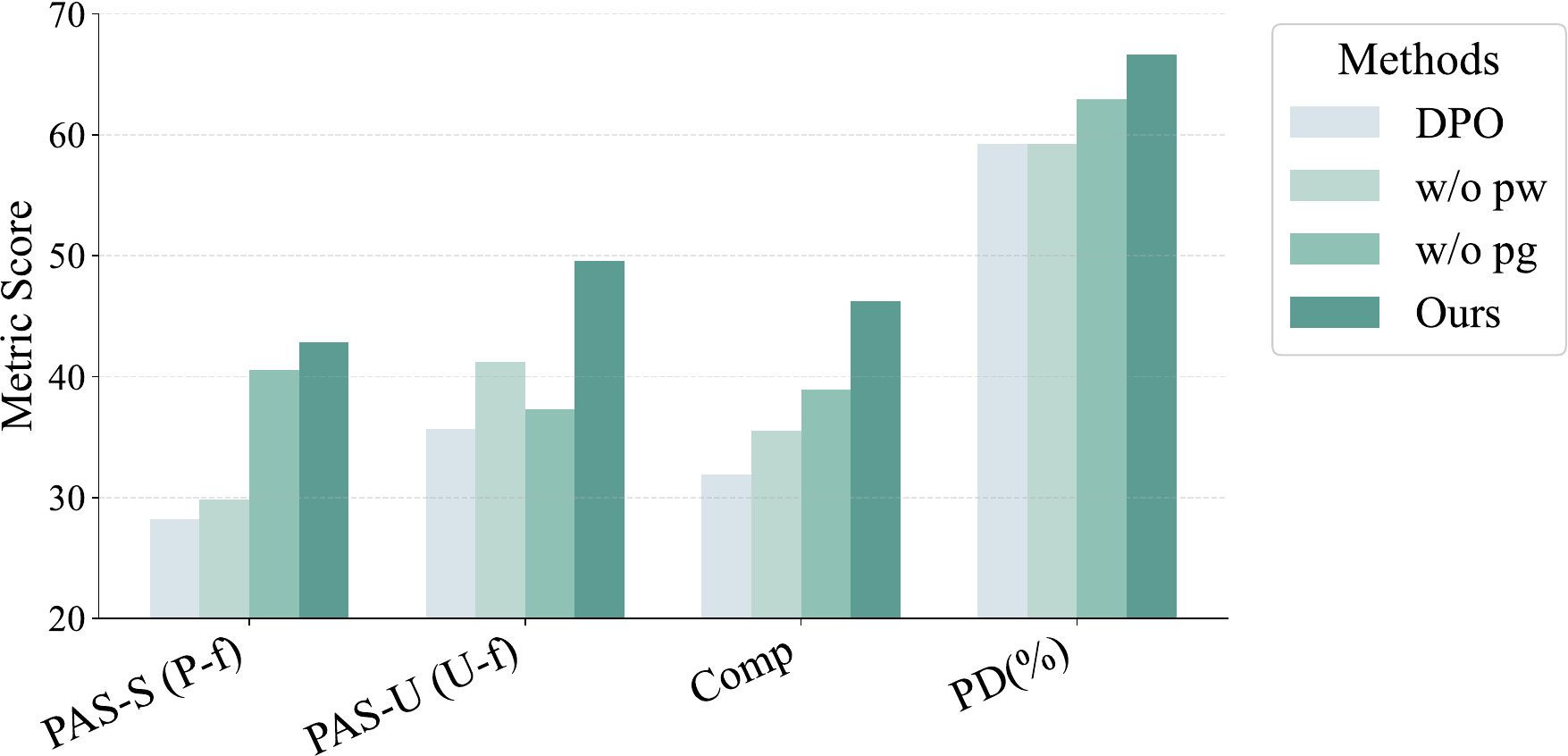}
    \caption{Ablation comparison on representative persona-related metrics, including PAS-S (P-F), PAS-U (U-F), Compliance, and PD. Removing either preference-intensity weighting or padding gating degrades performance, while the full TIPO consistently achieves the best overall results.}
    \label{fig:5_ablation}
\end{figure}
%

\subsection{Comparison Against Baselines}

We compare our TIPO with Frozen, SFT, DPO, ORPO, CPO, SimPO, and IPO. 
The results are shown in Table \ref{table1.tex} and Fig \ref{fig:4_radar}.
Overall, TIPO strikes the optimal balance between different dimensions.

In terms of \textbf{task performance}, TIPO achieves the best overall SR of 65.60\% among all methods, and obtains the best on the U-f with 62.11\% and the second best (-1.23\%) on P-f. 
These results indicate that TIPO does not significantly compromise task executability while performing persona alignment.
For \textbf{persona adherence}, TIPO outperforms the compared methods on multiple key metrics. Specifically, it achieves the highest PAS-S under Privacy-first (42.85) and the lowest PAS-S under Utility-first (15.71). 
It also attains the highest Compliance score of 46.22. These results suggest that TIPO is better able to generate behaviors consistent with the target persona while reducing actions that conflict with persona preferences. 
TIPO ranks second (-15.72) on PAS-U under Utility-first, because Frozen is less responsive to privacy-related preferences and therefore defaults more easily to utility-oriented behaviors.

Regarding \textbf{persona distinction}, TIPO achieves the best PD score of 66.67\%, outperforming all baselines. 
The result indicates that TIPO is more effective at generating execution trajectories with clearly different preference orientations under different persona conditions for the same task.

\subsection{Performance Across Task Categories}
To further analyze the performance of TIPO across different task scenarios, we group the original tasks into three categories according to the primary way in which privacy preference affects trajectory selection: Browsing \& Interaction (B \& I), Account \& File Management (A \& F), and Transactional Tasks (Trans). The results are shown in Table \ref{table3_single-column.tex}.

As shown in the results, TIPO exhibits relatively stable persona distinction ability across all three task categories, achieving PD scores of 66.67\%, 80.00\%, and 62.50\%, respectively. In Browsing \& Interaction tasks, the model achieves relatively high Compliance (53.31). In Account \& File Management tasks, it reaches 96.89 on PAS-S (Privacy-first) and obtains the highest PD score of 80.00\%. In Transactional Tasks, the model achieves the highest Overall SR of 72.17\%. Overall, these results suggest that TIPO generalizes well across different task categories.

Beyond the aggregate scores, the category-wise breakdown reveals that persona differences take distinct structural forms across task categories, which in turn leads to different metric patterns.

In \textbf{Browsing \& Interaction} tasks, persona differences are usually concentrated at a small number of localized and semantically explicit privacy decision points, rather than reshaping the whole trajectory. Typical examples include whether to enable incognito mode, whether to accept tracking, and whether to clear browsing traces after use. In these cases, the two personas still share most of the task backbone, while diverging only at a few clearly identifiable choices. This makes the preference signal relatively easy to localize and learn, since the model mainly needs to distinguish what should be done and what should be avoided at several key steps. As a result, this category tends to exhibit relatively high Compliance.

In \textbf{Account \& File Management} tasks, persona differences are usually more explicit and substantial. For example, Privacy-first users tend to prefer local saving or avoiding synchronization, whereas Utility-first users are more likely to choose cloud upload, sharing, or cross-device synchronization for greater convenience. These larger action-level differences make the two persona branches easier to distinguish, which naturally leads to higher persona distinction in this category.

By contrast, in \textbf{Transactional Tasks}, persona differences are more often expressed as soft trade-offs within a largely shared task flow, rather than as explicit local decisions. The two personas usually follow the same main purchase or booking process, and differ only at selected points, such as whether to accept recommendations, reuse stored information, or conduct additional cross-platform comparison. As a result, this category tends to maintain relatively high SR, while showing weaker persona distinction than categories with more explicit branching.

These results suggest that persona preferences influence different tasks in different ways, leading to various performance patterns. 
When the actions differences between personas are more explicit, the model can more easily separate the two trajectories and achieve stronger adherence and distinction, and vice versa. 

\begin{table}
\centering
\caption{Category-wise performance of TIPO on the Privacy Preference Dataset across Browsing \& Interaction, Account \& File Management, and Transactional Tasks.}
\label{table3_single-column.tex}
\begin{tabular}{cccccc} 
\hline
              & \textbf{SR}                                          & \textbf{PAS-S}                                     & \textbf{PAS-U}                                     & \textbf{Compliance}                                & \textbf{PD}                                           \\ 
\hline
\textbf{Type} & \textit{Overall}                                     & \textit{P-f}                                       & \textit{U-f}                                       & \textit{Comp}                                      & /                                                     \\ 
\hline
B \& I         & \textcolor[rgb]{0.659,0.804,0.776}{\textbf{59.85\%}} & \textbf{\textcolor[rgb]{0.659,0.804,0.776}{49.84}} & \textbf{\textcolor[rgb]{0.29,0.529,0.498}{56.78}}  & \textbf{\textcolor[rgb]{0.659,0.804,0.776}{53.31}} & \textbf{\textcolor[rgb]{0.659,0.804,0.776}{66.67\%}}  \\
A \& F         & 59.17\%                                              & \textbf{\textcolor[rgb]{0.29,0.529,0.498}{96.89}}  & 27.94                                              & \textbf{\textcolor[rgb]{0.29,0.529,0.498}{62.42}}  & \textbf{\textcolor[rgb]{0.29,0.529,0.498}{80.00\%}}   \\
Trans         & \textbf{\textcolor[rgb]{0.29,0.529,0.498}{72.17\%}}  & 23.33                                              & \textbf{\textcolor[rgb]{0.659,0.804,0.776}{51.47}} & 37.40                                              & 62.50\%                                               \\
\hline
\end{tabular}
\end{table}


\subsection{Ablation Study}

To verify the usefulness of each component in TIPO, we conduct ablation studies, and the results are shown in Table \ref{table2_single-column.tex} and Fig \ref{fig:5_ablation}. 
Overall, removing either module leads to performance degradation.

Specifically, after removing preference-intensity weighting, the model shows clear degradation in Overall SR, Compliance, and PD. After removing padding gating, the overall performance remains better than that of DPO, but falls short of the full model in terms of Compliance and PD. In contrast, the complete TIPO achieves the highest Compliance of 46.22\% and the highest PD of 66.67\%, indicating that the combination of the two modules yields the best overall performance.

These ablation results reveal that the gains of TIPO arise from the complementary effects of preference-intensity weighting and padding gating. The two components address different challenges in trajectory preference learning. On the one hand, persona-sensitive steps are often sparse within long trajectories and can easily be overwhelmed by a large number of neutral operations. Preference-intensity weighting explicitly amplifies these critical preference signals, enabling the model to focus more on decision points that truly reflect persona differences rather than treating all steps equally. On the other hand, variable-length trajectory alignment introduces placeholder steps such as \texttt{no\_action}, which bring additional noise into training. Padding gating suppresses invalid gradients introduced by such alignment artifacts, preventing the model from mistakenly learning placeholder patterns as preference signals.

Taken together, these results suggest that the two modules improve preference optimization from two complementary directions: effective signal enhancement and invalid noise suppression. For trajectory preference learning with structural heterogeneity, the uniform comparison scheme of standard DPO is therefore insufficient. A more effective solution must jointly address both the sparsity of persona-critical steps and the interference caused by alignment noise.

\section{Discussion}

\textbf{Extending TIPO beyond privacy.}
Although we study privacy preferences in this work, the TIPO framework is not limited to privacy itself. The key assumption behind TIPO is that, for the same task goal, different users may prefer different execution trajectories. Privacy is a representative and high-stakes example of this phenomenon, because it often leads to clear and observable differences in operation trajectories. However, the same formulation can also be applied to other user-specific factors that shape trajectory selection, such as efficiency preference, cost sensitivity, risk tolerance, or accessibility-related interaction needs. 
In this sense, privacy should be viewed as an entry point rather than the only adoption scenario. More broadly, our results suggest that trajectory-level preference alignment may provide a useful direction for building Mobile agents that adapt not only to what users want to achieve, but also to how they prefer the task to be carried out.

\textbf{Implications for agent developers.}
Our findings also have practical implications for the development of personalized Mobile GUI agents. In many mobile tasks, the task goal remains unchanged across users, while the preferred execution trajectory varies according to personal preferences. This makes trajectory-level preference modeling a practical addition to existing agent training pipelines. In particular, developers can adopt a framework such as TIPO by collecting paired trajectories under different user profiles for the same task and using them as preference supervision. This design does not require redefining the task objective, but instead focuses on aligning the generated trajectory with the target user preference. Such a setup is especially useful in scenarios where user trust depends not only on successful completion, but also on whether the agent follows an acceptable interaction process. Therefore, rather than treating personalization only as a matter of output customization, agent developers may also need to consider trajectory selection itself as an important part of personalized behavior.

\textbf{Limitations.}
TIPO mainly improves persona alignment at the trajectory level, rather than substantially improving the base Mobile GUI agent’s fundamental grounding, planning, or task execution ability. It should be viewed as a preference alignment framework built on top of an existing agent, rather than a general capability enhancement method.

\section{Conclusion}

In this paper, we studied personalized trajectory selection for Mobile GUI agents under privacy preferences. We showed that different personas can induce structurally heterogeneous trajectories, which makes standard preference optimization less effective in this setting. To address this issue, we proposed Trajectory Induced Preference Optimization (TIPO), which improves preference learning through preference-intensity weighting and padding gating. Experiments on the Privacy Preference dataset showed that TIPO improves persona adherence and persona distinction while preserving strong task executability. These results suggest that trajectory-level persona alignment is an important step toward more practical and user-aware Mobile GUI agents, with potential to support broader forms of personalization beyond privacy.

\section{Acknowledgments}
This work was supported in part by the Shandong Province Overseas Young Talents Program (2026HWYQ-009) and the Key Research and Development Program of Shandong Province  (2025CXGC010901).

\bibliographystyle{acmmm_template/ACM-Reference-Format}
\bibliography{main}

\end{document}